\documentclass[10pt,twocolumn,letterpaper]{article}

\usepackage{iccv}
\usepackage{times}
\usepackage{epsfig}
\usepackage{graphicx}
\usepackage{amsmath}
\usepackage{amssymb}
\usepackage{subcaption}
\usepackage{adjustbox}
\usepackage{multirow}
\usepackage[breaklinks=true,bookmarks=false]{hyperref}
\iccvfinalcopy
\def\iccvPaperID{****}

\begin{document}

\title{Benchmarking a Benchmark: How Reliable is MS-COCO?}

\author{
Eric Zimmermann\quad Justin Szeto\quad Jerome Pasquero\quad Frederic Ratle \\
Sama \\
Montréal, Québec \\
{\tt\small \{ezimmermann,jszeto,jpasquero,fratle\}@samasource.org}
}

\maketitle

\begin{abstract}
   Benchmark datasets are used to profile and compare algorithms across a variety of tasks, ranging from image classification to segmentation, and also play a large role in image pretraining algorithms. Emphasis is placed on results with little regard to the actual content within the dataset. It is important to question what kind of information is being learned from these datasets and what are the nuances and biases within them. In the following work, Sama-COCO, a re-annotation of MS-COCO, is used to discover potential biases by leveraging a shape analysis pipeline. A model is trained and evaluated on both datasets to examine the impact of different annotation conditions. Results demonstrate that annotation styles are important and that annotation pipelines should closely consider the task of interest. The dataset is made publicly available at \url{https://www.sama.com/sama-coco-dataset/}.
   
\end{abstract}

\section{Introduction}\footnotetext[1]{Accepted at ICCV 2023 DataComp Workshop}
Dataset benchmarks and evaluation standards play a pivotal role in shaping the direction and momentum of research in computer vision. They are the rulers against which the community measures the success and innovation of machine learning algorithms. These components are often times assumed to be monolithic artifacts that are acquired and analyzed to ensure reliability and quality for all algorithms thrown their way. Researchers and practitioners alike spend countless hours tuning their experiments in an attempt to maximize performance on said benchmarks, however, what happens when benchmarks themselves are less than ideal?

Vision datasets built for classification, detection, and segmentation are routinely used to benchmark algorithms or pretrain large neural networks. This is problematic as real world objectives are not always aligned with the data provided \cite{aln}. This misalignment may be correlated to the use of automatic label protocols and uncoordinated crowdsourcing efforts \cite{aln, aln-imgnet}. Alignment therefore requires a rigorous end-to-end pipeline where meticulously defined instructions based on a real-world task drives the annotation process.

The Common Objects in Context (MS-COCO) \cite{coco} is a standard benchmark used to evaluate and compare the quality of detection and instance segmentation algorithms which include but are not limited to YOLO \cite{YOLO}, R-CNN \cite{RCNN, FRCNN}, and DETR \cite{DETR}. It is composed of natural images, with applications in the autonomous driving industry, and as a result, sets the quality standard for neural networks developed on top of it. Due to the ubiquity of MS-COCO as a benchmark in computer vision, it is important to understand the reliability and quality of the bounding boxes and segmentation masks used to learn trends in data. In order to assess said quality, a replica of the dataset can be created to compare and find potential disagreements which may or may not be problematic. 

\begin{figure}[t]
    \centering
    \begin{subfigure}[b]{0.49\columnwidth}
    \centering
        \includegraphics[width=0.95\linewidth]{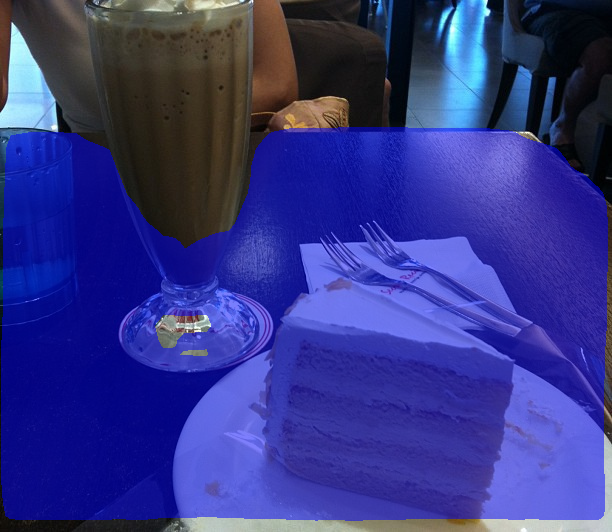}
    \end{subfigure}
    \begin{subfigure}[b]{0.49\columnwidth}
    \centering
        \includegraphics[width=0.95\linewidth]{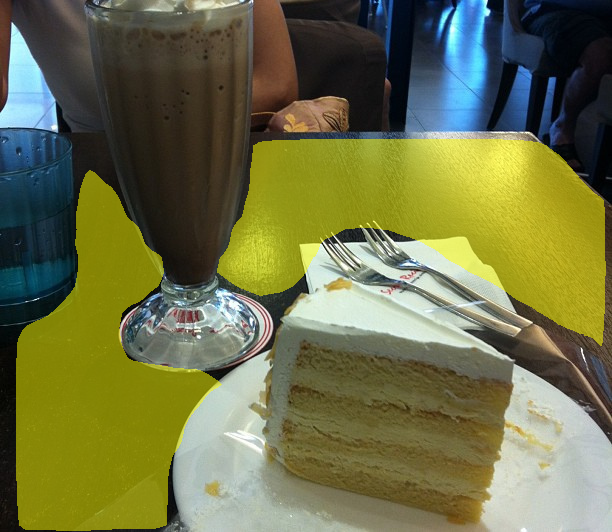}
    \end{subfigure}
    \caption{Model predictions when annotations do not go around occluders (blue) and go around occluders (yellow).}
    \label{fig:model-preds}
\end{figure}

In the following work, a re-annotation of MS-COCO is introduced, titled Sama-COCO, which places an emphasis on tight polygons that are true to pixel level labels as well as decomposed crowd instances. A Faster R-CNN is trained and instances across both variation of COCO are compared in order to detect and analyze differences in shapes as well as their impact on the features learned. Analysis shows that MS-COCO contains boundary noise and systematically avoids annotating around occluding objects, which leads to variations in the types of outputs produced by the model. 

\section{Dataset}

The Sama-COCO dataset represents a re-annotation effort of the existing MS-COCO dataset, spearheaded by a team of annotators. Initially conceived as an internal effort for generating high-quality ground truth data, the initiative evolved to offer a new lens through which to understand the intricate parameters that collectively define the quality of machine learning datasets.

The dataset was generated over a span of several months with a variable workforce: at times as many as 500 annotators worked simultaneously. A key focus involved meticulous guidelines for the annotators. Like the MS-COCO dataset, annotations are provided in the vector polygon format. 

Annotators were instructed to be extremely precise in drawing the polygons that identify the contours of COCO objects, minimizing the amount of background captured. They were also guided to give priority to annotating individual instances of objects rather than crowds. If an image contained more instances of a particular object class than a given threshold, annotators were directed to individually label the initial instances and categorize the remaining as a crowd. The prescribed threshold value varied throughout the project in order to balance considerations related to budget, time, and resulting data quality. Additionally, annotators were instructed to ignore objects measuring less than 10×10 pixels in size.

The re-labeling process incorporated all 123,287 training and validation images from the MS-COCO dataset. These images were preloaded with existing MS-COCO annotations, giving annotators the flexibility to modify, retain, or discard these as they saw fit. Quality assurance (QA) followed the annotation phase, where QA specialists inspected each submission. Submissions not meeting the quality criteria were sent back for refinement until they achieved the desired standard. Note that some annotators interpreted the instruction to ignore small objects as an instruction to delete the existing MS-COCO pre-annotations, whereas others simply left them untouched.

\begin{figure}[t]
    \centering
    \includegraphics[width=\columnwidth]{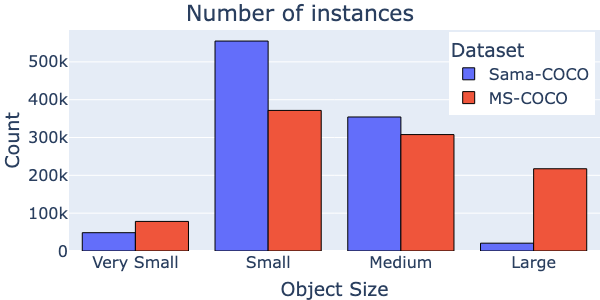}
    \caption{Object size distribution excluding crowd instances}
    \label{fig:object-sizes}
\end{figure}

In comparing Sama-COCO with the original MS-COCO dataset, several noteworthy differences surface. Firstly, the Sama-COCO dataset contains a significantly higher number of instances classified as crowds. This is partially attributable to the directive for annotators to deconstruct large singular crowds into smaller components and individual entities. Although both datasets share a common base, Sama-COCO has a greater number of instances for 47 out of the 80 classes. For certain classes, like \textit{person}, this increment is quite significant. Secondly, Sama-COCO exhibits almost double the number of vertices, a direct outcome of the instruction to maximize precision in polygon drawing. Furthermore, as depicted in Figure \ref{fig:object-sizes}, the number of large objects decreased considerably, as individual elements within large crowds or object clusters were re-labeled as distinct entities. Another key observation in the Sama-COCO dataset is the marked reduction in the number of very small objects (those with dimensions of 10×10 pixels or smaller compared to MS-COCO). Finally, one can also observe the emergence of a significantly greater number of small (ranging from 10×10 to 32×32 pixels) and medium-sized objects (ranging from 32×32 to 96×96 pixels) in the Sama-COCO dataset. 

\section{Shape Analysis}
Given that Sama-COCO is a re-annotation and not a correction of the initial dataset, there are no correspondences between annotations across samples. In order to confidently analyze differences in annotation shapes, polygons must be matched first. The analysis requirements are relaxed to single polygon shapes and leverage the notion of bounding box shape consistency. Shape consistency assumes that contour errors do not imply box errors. A match is determined using an overlap criterion based on the intersection over union (IoU) metric. For any pair of closed shapes $\mathbf{x}, \mathbf{y}$, IoU is defined as:

\begin{equation}
    \textrm{IoU}(\mathbf{x}, \mathbf{y}) = \frac{\vert\mathbf{x} \cap \mathbf{y}\vert}{\vert\mathbf{x} \cup \mathbf{y}\vert}.
\end{equation}

A match between two annotation instances across the datasets is defined by the pair of shapes with the highest IoU greater than a confidence threshold $T$ for all shapes in an image. Each annotation has at most one match, and there is no guarantee a match may be found. A matching threshold is empirically selected to be $0.90$. This strategy finds matches subject to contour noise and not related to global box error. For a shape $\mathbf{x}$ and a set of shapes $Y$ a match is defined as:

\begin{equation}
    \operatorname*{\max}_{\mathbf{y} \in Y} \textrm{IoU}(\mathbf{x}, \mathbf{y}), \qquad \textrm{IoU}(\mathbf{x}, \mathbf{y}) > T. 
\end{equation}

Once matches are found, differences are quantified between paired shapes using contour analysis. Let $(\mathbf{\partial{x}}, \mathbf{\partial{y}})$ denote the pair of contours for paired shapes $(\mathbf{x}, \mathbf{y})$ with lengths $(\Vert\mathbf{\partial{x}}\Vert, \Vert\mathbf{\partial{y}}\Vert)$. $D$ is denoted as the exact distance transform (EDT) \cite{exact_distance_transform} over a contour in the spatial domain $\Omega \subset \mathbb{R}^2$ where $\mathbf{p}$ defines the 
spatial location in $\Omega$. The average surface distance $d_{\mu}(\mathbf{x}, \mathbf{y})$ is the metric used to quantify average differences between shapes and is defined as:

\begin{equation}
    d_{\mu}(\mathbf{x}, \mathbf{y}) = \frac{\int_{\mathbf{\partial{x}}} D_\mathbf{\partial{y}}(\mathbf{p})\mathbf{dp} + \int_{\mathbf{\partial{y}}} D_\mathbf{\partial{x}}(\mathbf{p})\mathbf{dp} }{\Vert\mathbf{\partial{x}}\Vert + \Vert\mathbf{\partial{y}}\Vert}.  
\end{equation}

Some paired shapes may have small sub-regions that have large disagreements. In this scenario, the average surface distance is not capable of capturing this phenomenon. To mitigate this issue, the maximum distance $d_{\max}(\mathbf{x}, \mathbf{y})$ is introduced and defined as:

\begin{equation}
    d_{\max}(\mathbf{x}, \mathbf{y}) = \max(\operatorname*{\max}_{\mathbf{p} \in \mathbf{y}} D_\mathbf{\partial{x}}(\mathbf{p}), \operatorname*{\max}_{\mathbf{p} \in \mathbf{x}} D_\mathbf{\partial{y}}(\mathbf{p})).
\end{equation}

\begin{figure}[t]
\centering
    \includegraphics[width=\columnwidth]{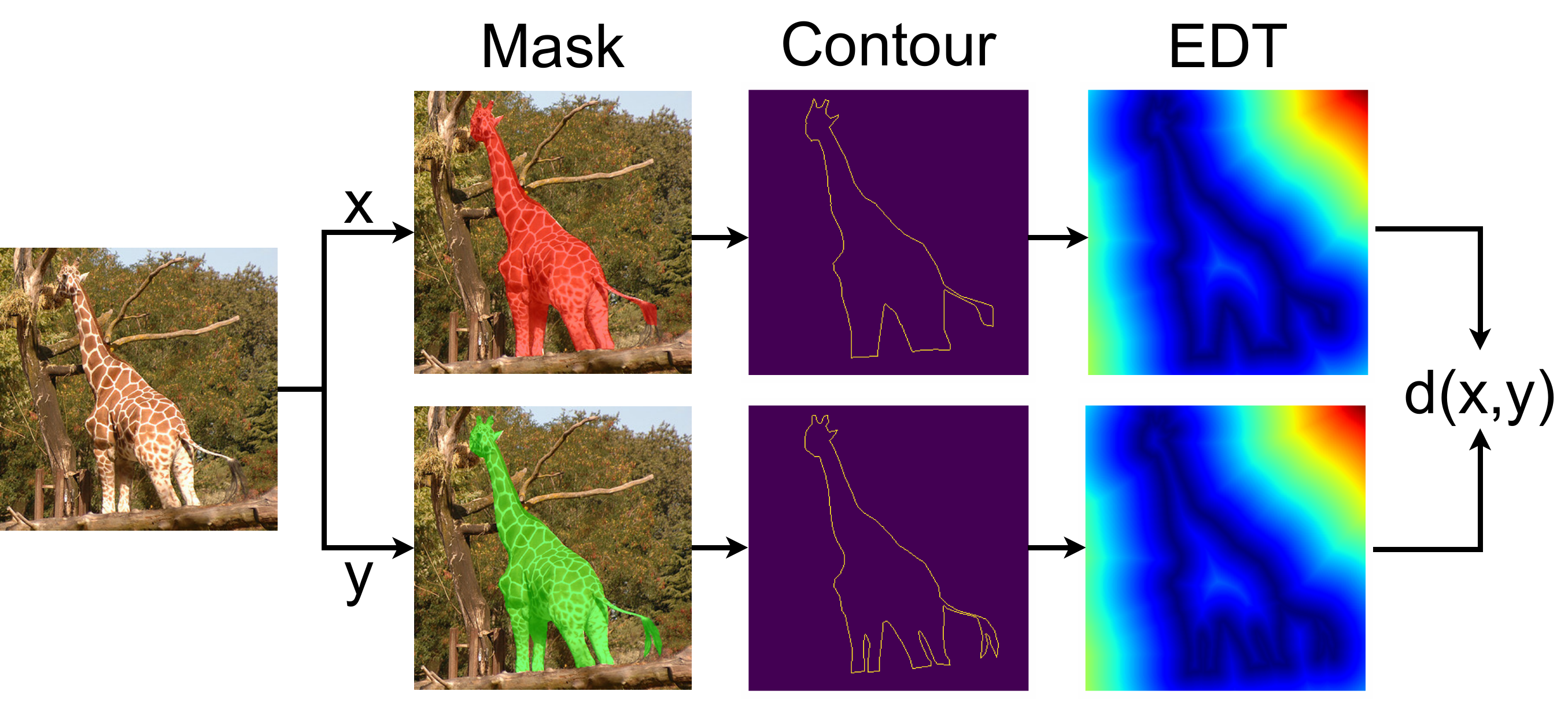}
     \caption{Surface distance pair matching pipeline}
    \label{fig:match-pair}
\end{figure}

\begin{figure}[h]
    \centering
    \begin{subfigure}[b]{\columnwidth}
        \includegraphics[width=0.95\linewidth]{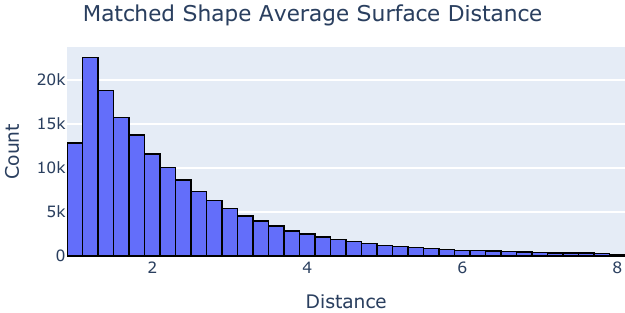}
    \end{subfigure}
    \begin{subfigure}[b]{\columnwidth}
        \includegraphics[width=0.95\linewidth]{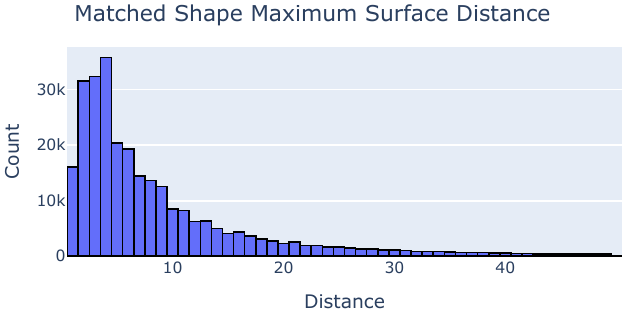}
    \end{subfigure}
    \caption{Long-tail distribution of average and maximum surface distance with values greater than one pixel and up to three standard deviations.}
    \label{fig:surface-dists}
\end{figure}

The matching pipeline is applied to the training split and $310504$ confident matches are found. Each shape is rasterized to a mask using pycoco standards, and contours are generated by subtracting the binary erosion of the mask with itself. The EDT is generated and the path integral is computed by indexing the distance map with the contour of the paired shape. This pipeline is done bidirectionally for both shapes and is visualized in Figure \ref{fig:match-pair}. The distribution of average and maximum surface distances are visualized in Figure \ref{fig:surface-dists}.

\section{Experiments}

The impacts of the re-annotation process consisting of tighter polygons, broken up crowds, and more annotation instances is studied by training and evaluating the quality of a neural network's predictions on detection and instance segmentation tasks. A Faster R-CNN with a ResNet-50 and FPN backbone is trained on both MS-COCO and Sama-COCO using the Detectron2 framework \cite{detectron2, resnet, fpn} and evaluated using the standard MS-COCO benchmark, treating each dataset's validation split as ground truth. The neural network is trained with automatic mixed precision for a total of $270k$ iterations with a batch size of $16$ across $8$ Nvidia V100 GPU's.  All hyperparameters held constant across all experiments and are evaluated using mean average precision (mAP) as seen in Table \ref{tab:train-results}.

\begin{table*}
\centering
    \begin{tabular}{|l|l|l|c|c|c|c|c|}
        \hline
        Task &Training Set & Validation Set & mAP & mAP@50 & mAP Large & mAP Medium & mAP Small \\
        \hline
        \hline
        
        \multirow{4}{*}{Detection} &MS-COCO & MS-COCO & 40.260 & 61.108& 51.977 & 43.550 & 24.298\\
        &Sama-COCO & MS-COCO & 38.353 & 58.793 & 50.468 & 41.785 & 22.857\\
        &MS-COCO & Sama-COCO & 39.092 & 58.396 & 53.546 & 42.981 & 22.918\\
        &Sama-COCO & Sama-COCO & 40.859 & 60.817 & 54.546 & 44.845 & 25.741\\
        \hline
        \multirow{4}{*}{Segmentation}&MS-COCO & MS-COCO & 37.193 & 58.662 &53.297 & 39.526 & 18.658\\
        &Sama-COCO & MS-COCO & 34.916 &	56.652 & 51.222 & 37.755 & 16.961\\
        &MS-COCO & Sama-COCO & 36.523 & 56.283 & 55.457 & 40.134 & 17.521\\
        &Sama-COCO & Sama-COCO & 38.925 & 59.506 & 57.221& 42.551 & 21.299\\
        \hline
    \end{tabular}
    \caption{Detection and segmentation results}
    \label{tab:train-results}
\end{table*}

We evaluate the theoretical implications of learning perfect representations that match the validation set. In this scenario, we compare source annotations to target annotations, treating the source as the model predictions and the targets as ground truth. We alternate source and targets to be both MS-COCO and Sama-COCO  to ensure evaluation is fair. Results are displayed in Table \ref{tab:ann-results}.

\begin{table*}
\centering
\begin{tabular}{|l|l|l|c|c|c|c|c|}
    \hline
    Task & Source & Target & mAP & mAP@50 & mAP Large & mAP Medium & mAP Small \\
    \hline
    \hline
    \multirow{2}{*}{Detection} & MS-COCO & Sama-COCO & 46.968 & 62.417 & 65.421 & 51.458 & 33.007 \\
    &Sama-COCO & MS-COCO & 48.298 & 64.086 & 63.562 & 52.237 & 35.604\\
    \hline
    \multirow{2}{*}{Segmentation}&MS-COCO & Sama-COCO & 41.032 & 62.061 & 66.570 & 45.849 & 25.441 \\
    &Sama-COCO & MS-COCO & 41.744 & 62.906 & 64.826 & 47.018 & 26.433\\
    \hline
\end{tabular}
\caption{Detection and segmentation results when treating source dataset as model predictions against target dataset}
\label{tab:ann-results}
\end{table*}

\section{Discussion}

It is important to preface that no dataset is perfect and that Sama-COCO is not better or worse than MS-COCO. Every dataset will always contain unavoidable bias, however, different forms of bias play different roles in how they may affect a neural network's performance. This is observed by benchmarking each dataset with one another. When comparing matched instances between datasets, it is possible to observe systemic biases present in the MS-COCO dataset. These biases come in two different forms. 

The first form of bias is in relation to the tightness of the polygons around each instance. It is observed that paired polygons with low average surface distance differ subtly along the contour. On average, Sama-COCO polygons are tighter than the original annotations, however, the mix of under and over segmented instances may not play a role in true prediction quality if the noise is zero in expectation. It may also be possible that as networks scale up, they learn to fit to the bias in these contours, cheating the evaluation metric altogether. In this scenario, it becomes hard to discern between the true quality of the representations learned by a neural network, as the only means of evaluating them contain bias.

Secondly, pairs found with medium levels of average and maximum surface distance correspond to instances where occluders and annotation style guidelines are handled and specified differently. Sama-COCO placed an emphasis on having contours that are true to the pixels belonging to an observation, while the original dataset contains contours that circumvent these occluders altogether. Accounting for occluders is true to a pixel-level instance segmentation task, while avoiding them is more akin to a localization task. The neural network trained on this hidden bias will learn to solve these tasks differently. It is therefore imperative that any machine learning practitioner understands how an individual dataset is correlated to the downstream task they are trying to solve, and should place a large emphasis on annotation standards and guidelines during the data acquisition phase in order to mitigate issues from the top. It may not be wise to merge datasets with conflicting annotation style, since the downstream behavior of a neural network may not be predictable.

\begin{figure}[h]
    \centering
    \hfill
    \begin{subfigure}[b]{0.24\columnwidth}
        \includegraphics[width=0.95\linewidth]{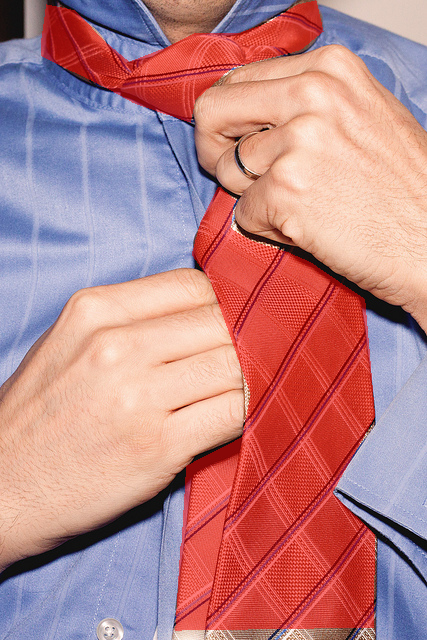}
    \end{subfigure}
    \hfill
    \begin{subfigure}[b]{0.24\columnwidth}
        \includegraphics[width=0.95\linewidth]{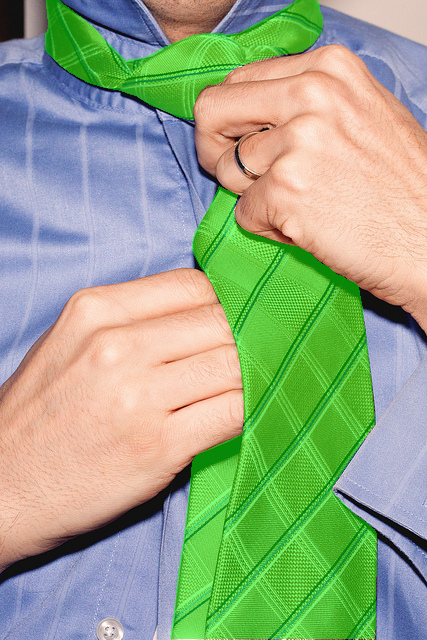}
    \end{subfigure}
    \hfill
    \begin{subfigure}[b]{0.24\columnwidth}
        \includegraphics[width=0.95\linewidth]{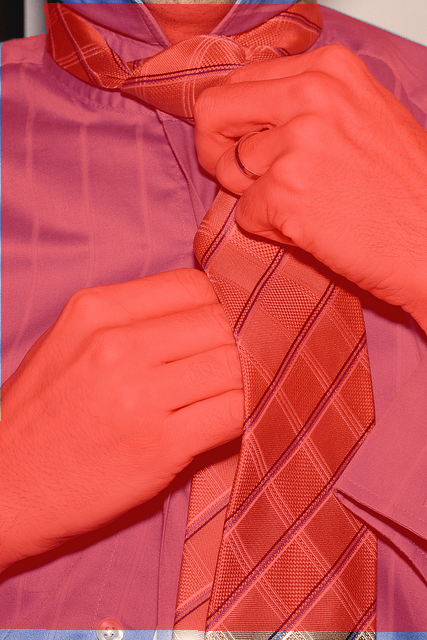}
    \end{subfigure}
    \hfill
    \begin{subfigure}[b]{0.24\columnwidth}
        \includegraphics[width=0.95\linewidth]{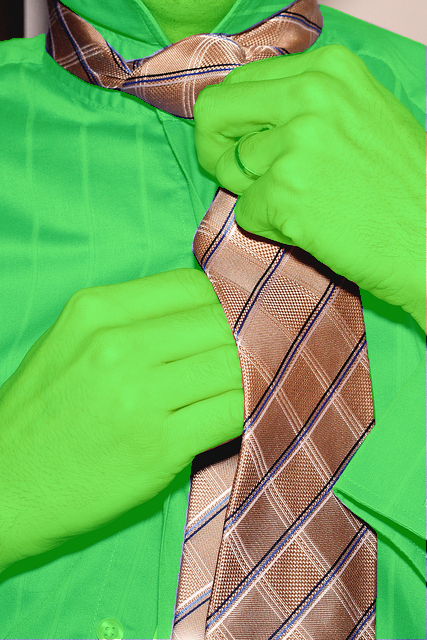}
    \end{subfigure}
    \hfill
    \caption{Annotation style difference with occluders. Both MS-COCO (red) and Sama-COCO (green) treat the person as an occluder to the tie. However, MS-COCO does not treat the tie as an occluder to the person.}
    \label{fig:ann-styles}
\end{figure}

When looking at the differences in evaluation metrics across detection and segmentation tasks, it is clear that networks benefit from evaluation on the same style of annotation as seen in Table \ref{tab:train-results}. This implies that performance is tightly connected to a subjective definition of quality. If you enrich a dataset with additional samples, contrary to the expectation, network performance may degrade if the style distribution changes.  This is validated theoretically when using validation annotations from one dataset against another. Even a perfect predictor on another dataset will suffer from missed instances, boundary deformations, and stylistic nuances. It is also noted that some state-of-the-art detection algorithms outperform these results \cite{codetr}. This is of interest because box annotations should be relatively constant regardless of polygon variations. This implies that a network is likely to overfit on the exact type of information from the training dataset that might not be replicable in another dataset.

\section{Conclusion}

It is clear from the discussion that bias in datasets can lead to unwanted or unexpected behavior that may be problematic. In the case of instance segmentation, the choices in annotation style affect the model's outputs for occluded objects. As such, careful consideration must be made when building annotated datasets to ensure that reflect the behavior in real world application. While Sama-COCO is not free from all labeling errors, it does provide a high-quality set of annotations that can be used to better explore the field of label noise and applications where tight polygons are essential. 

{\small
\bibliographystyle{ieee_fullname}
\bibliography{egbib}
}

\end{document}